\documentclass[runningheads,a4paper]{llncs}
\usepackage{amsmath,graphicx}
\usepackage{amssymb}
\usepackage{graphicx}
\usepackage{color}
\usepackage{url}
\usepackage[table]{xcolor}
\usepackage{multirow}
\usepackage{subfig}

\newcommand\blfootnote[1]{%
	\begingroup
	\renewcommand\thefootnote{}\footnote{#1}%
	\addtocounter{footnote}{-1}%
	\endgroup
}

\newcommand{\keywords}[1]{\par\addvspace\baselineskip
	\noindent\keywordname\enspace\ignorespaces#1}

\usepackage{array}
\newcolumntype{L}[1]{>{\raggedright\let\newline\\\arraybackslash\hspace{0pt}}m{#1}}
\newcolumntype{C}[1]{>{\centering\let\newline\\\arraybackslash\hspace{0pt}}m{#1}}
\newcolumntype{R}[1]{>{\raggedleft\let\newline\\\arraybackslash\hspace{0pt}}m{#1}}

\begin{document}

\title{Automatic Segmentation of the Left Ventricle in Cardiac CT Angiography Using Convolutional Neural Network}

\titlerunning{Convolutional Neural Networks for Cardiac CT Segmentation}

\author{Majd Zreik$^{1}$, Tim Leiner$^{2}$, Bob D.\,de Vos$^{1}$,\\ Robbert W. van Hamersvelt$^{2}$, Max A. Viergever$^{1}$, Ivana I\v sgum$^{1}$}

\institute{$^{1}$ Image Sciences Institute, University Medical Center Utrecht, The Netherlands \\     $^{2}$ Department of Radiology, University Medical Center Utrecht, The Netherlands }

\authorrunning{M. Zreik et al.}

%
\maketitle 
\begin{abstract}
Accurate delineation of the left ventricle (LV) is an important step in evaluation of cardiac function. In this paper, we present an automatic method for segmentation of the LV in cardiac CT angiography (CCTA) scans. Segmentation is performed in two stages. First, a bounding box around the LV is detected using a combination of three convolutional neural networks (CNNs). Subsequently, to obtain the segmentation of the LV, voxel classification is performed within the defined bounding box using a CNN. The study included CCTA scans of sixty patients, fifty scans were used to train the CNNs for the LV localization, five scans were used to train LV segmentation and the remaining five scans were used for testing the method.
Automatic segmentation resulted in the average Dice coefficient of 0.85 and mean absolute surface distance of 1.1 mm. The results demonstrate that automatic segmentation of the LV in CCTA scans using voxel classification with convolutional neural networks is feasible.

\keywords
Left ventricle segmentation, Convolutional Neural Network, Cardiac CT Angiography, Classification, Deep learning
\end{abstract}

\blfootnote{This work has been published as: Zreik, M., Leiner, T., de Vos, B. D., van Hamersvelt, R. W., Viergever, M. A., I\v{s}gum, I. (2016, April). Automatic segmentation of the left ventricle in cardiac CT angiography using convolutional neural networks. In Biomedical Imaging (ISBI), 2016 IEEE 13th International Symposium on (pp. 40-43). IEEE.}

\section{Introduction}
\label{sec:intro}

Cardiovascular disease (CVD) is the leading cause of death in the developed countries \cite{WHO14}. Segmentation of the left ventricle (LV) plays a fundamental step in the evaluation of cardiac function and thereby CVD \cite{Kang12}. In clinic, cardiac function is standardly evaluated with cardiac MR, but recent literature suggests that this can also be performed using cardiac CT angiography (CCTA) \cite{Tech11}. Therefore, segmentation of the LV and thereby assessment of cardiac function with CCTA has become topic of intensive research {\cite{Tech11,Xion15}. Manual delineation of the LV is a labor intensive task that is prone to intersubject variability \cite{Suge06}.

Several automatic methods for segmentation of the LV in CCTA have been proposed \cite{Tava13}. Different approaches have been described, including an atlas-based, boundary-driven, model-based and machine learning approaches.
Kiri{\c{s}}li et al. \cite{Kiri10} registered multiple atlases to a target CCTA image and propagated atlas labels to the target image. To obtain the segmentation of the four heart chambers, the propagated labels were merged using a per voxel majority voting. 
Marie-Pierre \cite{MP06} presented a method for segmentation  of the LV in CCTA images using graph cuts and active contours to find the LV boundary after global localization of the blood pool within the left ventricular cavity. Zheng et al. \cite{Zhen08a} presented a four-chamber heart segmentation in CCTA scans using marginal space learning and steerable features to locate the heart and to delineate its boundaries. Recently, Xiong et al. \cite{Xion15} used five anatomical landmarks to initialize an LV model, which was adapted to a target CCTA image by deformation increments guided by trained AdaBoost classifiers.

In this work, segmentation of the LV from CCTA scans using Convolutional Neural Network (CNN) is proposed. CNN is a machine-learning technique, which is increasingly becoming popular in image analysis \cite{Lecu15}. In contrast to traditional machine-learning methods that require careful feature engineering, CNN is an end-to-end technique; it requires image data as input and features needed for the classification are determined automatically. CNNs are widely used for processing of natural images \cite{Lecu15} and since recently also in medical image analysis (e.g. \cite{Wolt15}, \cite{Ciom15}).
In this work, segmentation is performed in two stages. First, the LV is localized by a combination of three CNNs, each detecting presence of the LV in all image slices of an image plane, creating a bounding box around it. Thereafter, a dedicated CNN was designed to identify voxels that belong to the LV by analyzing the volume within the bounding box.

\begin{figure}[h!]
	
	\begin{minipage}[b]{1.0\linewidth}
		\centering
		\centerline{\includegraphics[width=1.0\linewidth]{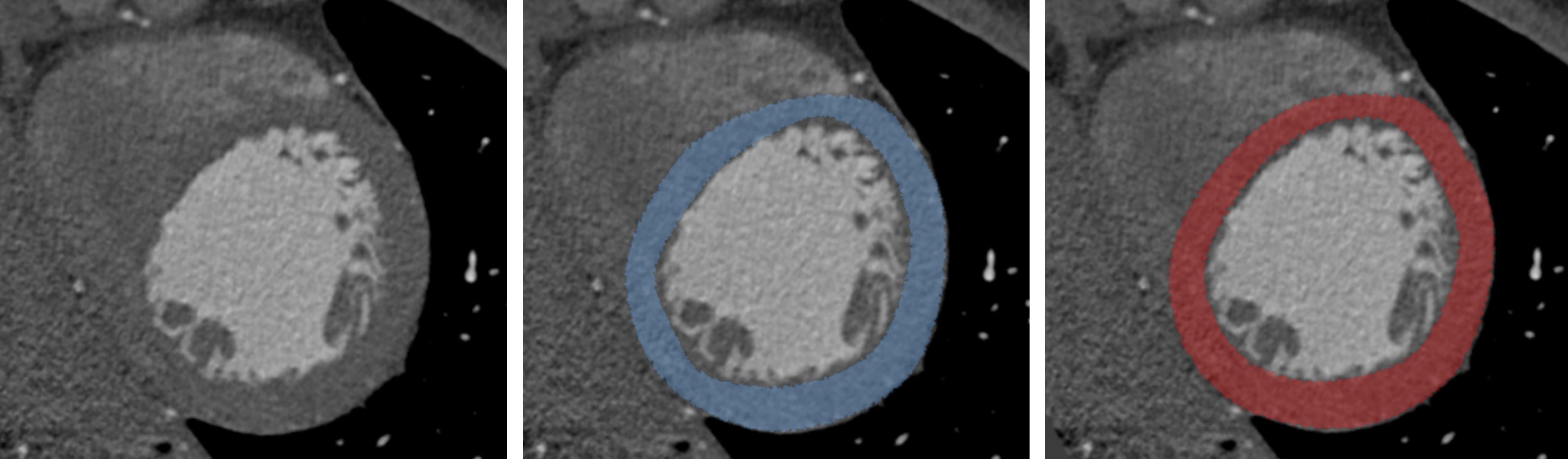}}
		
	\end{minipage}
	\caption{LV in one slice of a CCTA shown in short axis view (left), reference annotation (middle) and second observer's annotation (right).}
	\label{fig:obs1_obs2}
\end{figure}

\begin{figure*}[t!]
	
	\begin{minipage}[b]{1.0\linewidth}
		\centering
		\centerline{\includegraphics[width=1.0\linewidth]{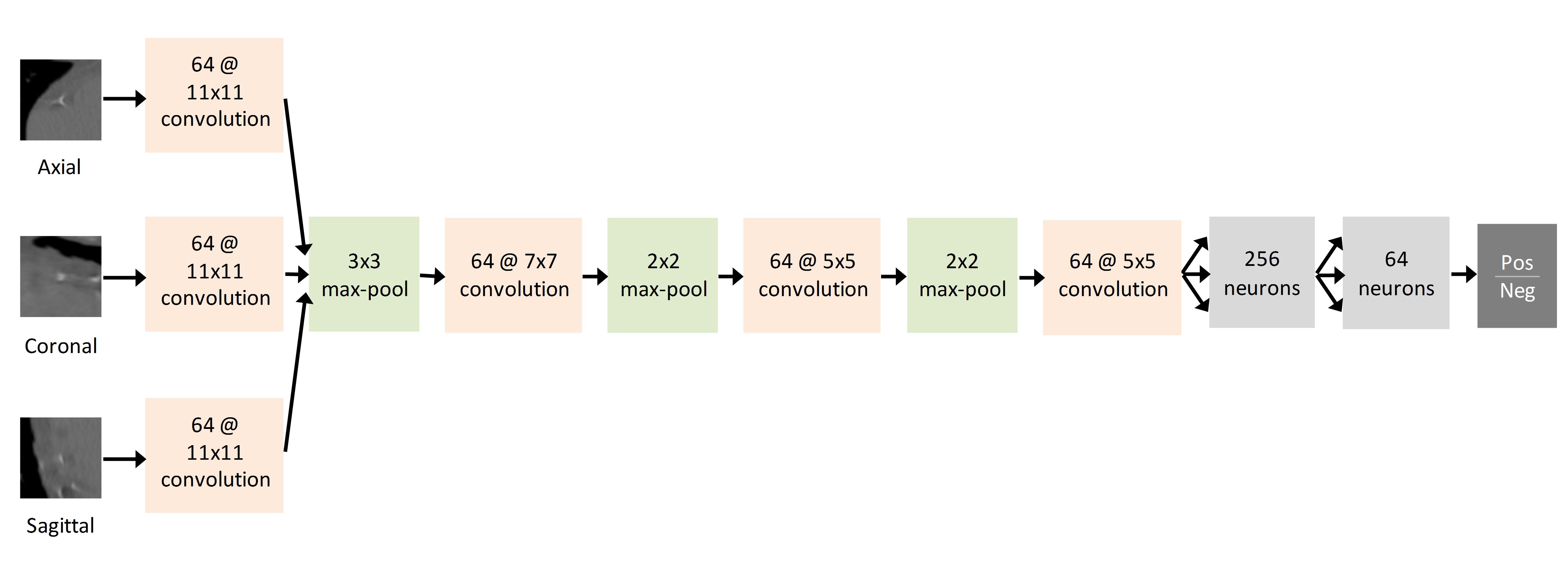}}
		
	\end{minipage}
	\caption{CNN architecture. The CNN has 4 convolutional layers, 3 max pooling layers, two fully connected layers and one softmax output layer. The input consists of three $48\times48$  patches from axial, sagittal and coronal image slices centered around the target voxel.}
	\label{fig:network}
\end{figure*}

\section{MATERIALS AND METHODS}
\label{sec:pagestyle}

\subsection{Data}
\label{ssec:data}

This study included retrospectively collected CCTA scans of sixty patients randomly divided in three sets: Fifty scans were used to train the CNNs for localization of the LV, five scans were used to train LV segmentation and the remaining five scans for testing the method. All scans were acquired on a 256-detector row scanner (Philips Brilliance iCT, Philips Medical, Best, The Netherlands) using 120 kVp and 210-300 mAs, with ECG-triggering and contrast enhancement.
Scans were reconstructed to 0.45 mm thick slices with 0.9 mm spacing and in-plane resolution of 0.38-0.49 mm. 

Reference volumes of interest for LV localization were delineated by a trained observer, who manually drew rectangular bounding boxes around the LV in each CCTA scan.

Reference standard for LV segmentation was defined by manual annotations of the LV using MeVisLab\footnote{http://www.mevislab.de} platform. Following clinical protocol, annotations were performed in short axis view of the heart, excluding myocardial fat, papillary muscles and the trabeculae carneae. In the manually determined short axis view, points were manually placed along the endocardium and epicardium in each third slice. From the defined points, closed contours for the endocardium and the epicardium were defined by cubic spline interpolation. The contours were propagated to the adjacent slices where they could be adjusted manually by moving existing or placing new points. Reference LV volume was defined as all voxels enclosed by the manually annotated endocardial and epicardial contours. 

Manual LV segmentation was performed by two trained observers. The first observer annotated all five training and five test scans which were used as the reference standard. To estimate interobserver agreement, a second observer, blinded to the results of the first observer, annotated five test scans. Figure \ref{fig:obs1_obs2} illustrates manual reference and second observer LV annotations.

\subsection{Segmentation using Convolutional Neural Networks}
\label{ssec:cnn}
LV segmentation was performed in two stages. First, to limit the analysis to the volume around the LV, a bounding box around the LV is defined. Second, to obtain LV segmentation, voxel classification is performed within the defined bounding box.

The bounding box is defined following the method proposed by de Vos et al. \cite{Vos16}. In brief, the method estimates the position of the LV using three AlexNet CNNs \cite{Kriz12} that determine the presence of the LV independently in axial, coronal and sagittal slices of the image volume. The combination of these per-slice probabilities yields a 3D bounding box around the LV. All voxels within the detected box are treated as candidates for the segmentation of the LV.

Voxels inside the bounding box are classified using a CNN illustrated in Figure \ref{fig:network}. The CNN architecture and its parameters were determined in a
pilot study, using leave-one-scan-out experiments on training
scans only. Three $48\times48$ pixel patches extracted around each voxel from axial, coronal and sagittal slices are used as input to the CNN. The patch size was chosen to be large enough to contain spatial anatomical information, while remaining small enough to keep the computational load limited. The CNN has 4 convolutional, 3 max-pooling and 2 fully connected layers. All units use rectified linear unit (ReLU) as an activation function \cite{Glor11a}. The dropout strategy \cite{Sriv14} was employed for all hidden units in the fully connected layers to prevent over-fitting and was set to 0.5. The output layer uses a softmax classifier to return a probability for each voxel to belong to the LV. 

To obtain a binary segmentation, volume with resulting posterior probabilities was smoothed by a 3D Gaussian filter with a kernel size of 1.5 mm and thereafter, thresholded at 0.4. To ensure that no single isolated voxels or small clusters of voxels are contained in the result, only the largest 3D-connected component is preserved.

\section{EXPERIMENTS AND RESULTS}
\label{sec:expres}

CNNs for LV localization were trained with all axial, coronal and sagittal slices from all training images for 30 epochs. Namely, 10,208, 11,191 and 9,848 positive and 15,392, 14,409 and 4,901 negative axial, coronal and sagittal slices were used, respectively. Visual evaluation of the automatic bounding box detection revealed that complete LV was contained within the box in all test scans.

Training the CNN for the LV segmentation was performed using patches around 100,000 voxels extracted from the LV (positive) and 100,000 other candidate voxels (negative) randomly sampled within the automatically obtained bounding box in each training image. The network was trained in 70 epochs.
Testing was performed using all voxels within the automatically determined bounding box around the LV.

Classification performance was evaluated using sensitivity and specificity, and on average, the method achieved sensitivity of 95.0\% and specificity of 96.6\% per scan.

Segmentation performance was evaluated using Dice coefficient as an overlap measure between reference and automatically segmented volume, and the mean absolute surface distance (MAD) between the automatically and manually annotated LV boundaries. Additionally, manual segmentations of the second observer were compared to the reference annotations using the same evaluation criteria. 
On average, the automatic method achieved a Dice coefficient of 0.85 and a MAD of 1.1 mm per scan. For the second observer, these were 0.9 and 0.6 mm, respectively. Table \ref{list:perf} lists these results. Figure~\ref{fig:exp} shows segmentation results in the short axis view for three different examples.

\begin{table}[h]
\centering
	\begin{tabular}{l*{6}{c}r}
		
		Measure              & Sens.[\%]& Spec.[\%]& Dice & MAD  \\
		\hline
		
		Test Set            & 95.0 & 96.6 & 0.85 & 1.1 mm   \\
		Second observer 	& - & - & 0.90 & 0.6 mm  \\

\end{tabular}
	\caption{Average sensitivity (Sens.), specificity (Spec.), Dice coefficient (Dice) and mean absolute distance (MAD) for the automatic method and the second observer over five test scans.}
	
	\label{list:perf}	
	
\end{table}

\begin{figure}[h!]
	\begin{minipage}[b]{1.0\linewidth}
		
		\centering{\includegraphics[width=1.0\linewidth]{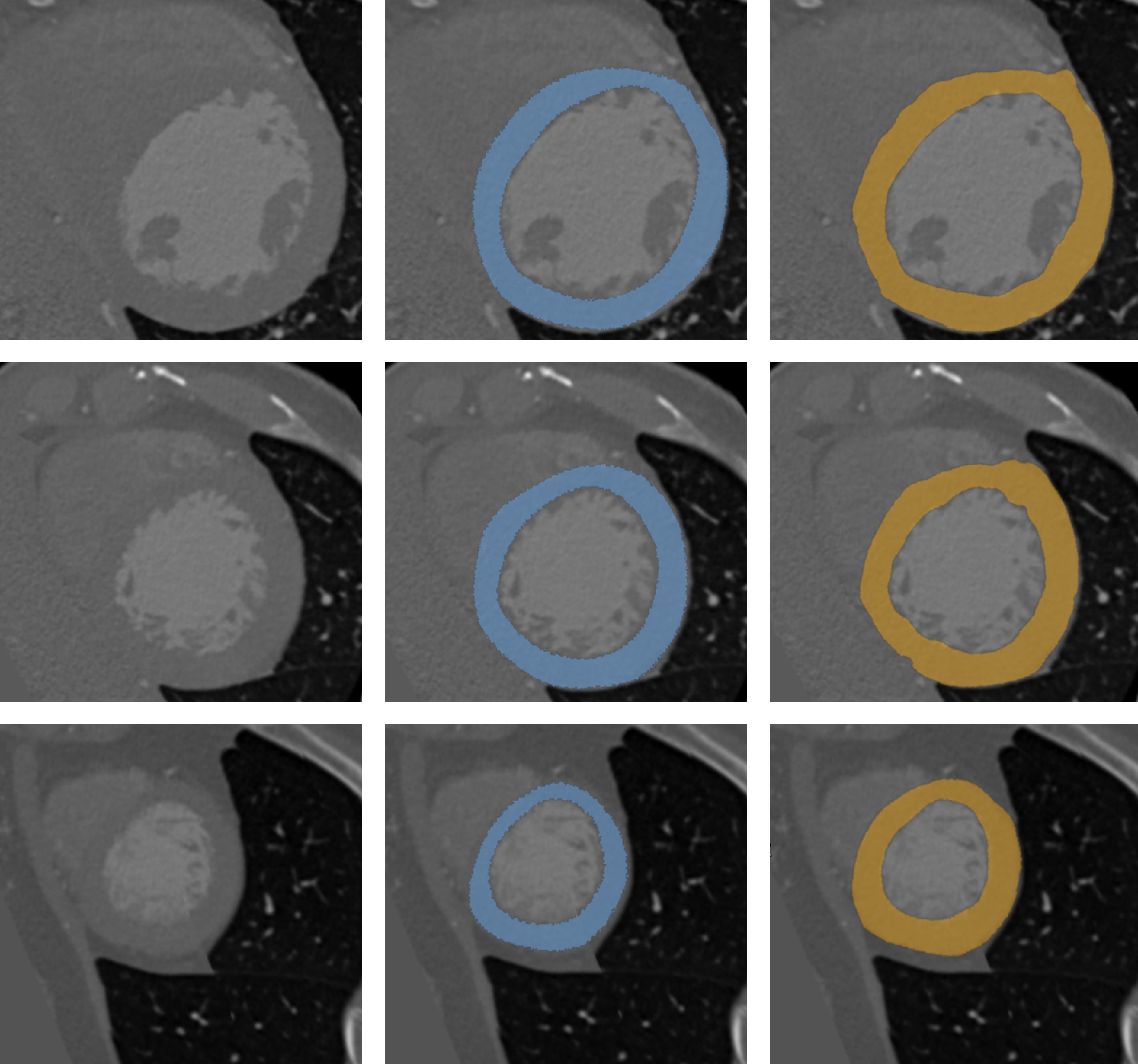}}
		
	\end{minipage}
	\caption{Segmentation results in three slices in short axis view of three different patients. Each column illustrates a slice from CCTA (left), corresponding manual reference annotation (middle) and  automatic segmentation (right).}
	\label{fig:exp}
\end{figure}

\section{DISCUSSION AND CONCLUSION}
\label{sec:dis}
An automatic method for segmentation of the LV in CCTA using CNNs has been presented. The method is based on voxel classification within an automatically defined bounding box around the LV. The box is defined using three independent CNNs, each detecting presence of the LV in all image slices of an image plane. Voxel classification is performed using a single CNN fed with three orthogonal image patches. To the best of our knowledge, this work is the first employing CNNs for LV segmentation.

The proposed method shows high sensitivity and specificity, high overlap and low distance between the automatic segmentation and the manual annotations. Even though the second observer outperforms the automatic method, visual inspection of the errors revealed similar disagreements along most of the LV boundary. 
Most of the false positives occurred due to absence of contrast between LV and surrounding cardiac tissue. Namely, the automatic method occasionally presents irregular boundary or relatively distant false positive errors, as can been seen in Figure~\ref{fig:exp}, top right. While experts use prior knowledge about the shape of the LV, the proposed method relies on texture analysis only. Future work will investigate whether integrating shape constraints with the described CNN might be beneficial.     

Moreover, several previous studies discarded apical slices from their performance evaluation due to poor performance or unreliable reference standard in that region \cite{Tava13}. In this work, evaluation has been performed on the complete scans. Visual inspection revealed similar performance in the apex as in other slices containing LV. 
Furthermore, visual evaluation showed that papillary muscles and trabeculae were accurately excluded from the automatic segmentation. Even though the intensity of these areas is similar to the intensity of the LV, the size of the patches used in voxel classification allowed this differentiation.

Previous publications typically reported performance using distance metric. However, in previous work, different data sets have been used and evaluation might have been performed using different implementation of the metric, hence comparison with other reported results has to be taken with caution. Xiong et al. \cite{Xion15} reported average distance to LV boundary of 2.79 mm, and Zheng et al. \cite{Zhen08a} reported distances of 1.13 mm and 1.21 mm to endo- and epicardium boundary, respectively. Both methods were evaluated using point to mesh distances. Furthermore, Kiri{\c{s}}li et al. \cite{Kiri10} evaluated segmentation on the complete surface and reported distances of 1.04 mm and 0.6 mm on epicarial and endocardial boundary, respectively. This demonstrates that our method achieved comparable performance. 

Several earlier publications performed segmentation of the LV as a part of four-chamber heart segmentation method. Given that the described method takes only image patches as input, the method might be straightforwardly extended to segmentation of four heart chambers in CCTA. 

In this work a small set of scans acquired with the same scanner, same acquisition protocol and without visible pathology was used to train the CNN for LV segmentation. In future work, we will extend the training and evaluation to a larger set of scans acquired across different scanner vendors and containing pathology.
 
To conclude, automatic segmentation of the LV in CCTA scans using convolutional neural networks is feasible.

\section{Acknowledgment}
We gratefully acknowledge the support of NVIDIA Corporation with the donation of the Tesla K40 GPU used
for this research.

\bibliographystyle{splncs03}
\bibliography{D:/Literature/CAD}

\begin{thebibliography}{10}
\def\url#1{}
\providecommand{\urlprefix}{URL }

\bibitem{Ciom15}
Ciompi, F., de~Hoop, B., van Riel, S.J., Chung, K., Scholten, E.T., Oudkerk,
  M., de~Jong, P.A., Prokop, M., van Ginneken, B.: Automatic classification of
  pulmonary peri-fissural nodules in computed tomography using an ensemble of
  {2D} views and a convolutional neural network out-of-the-box. Medical Image
  Analysis  (2015)

\bibitem{Glor11a}
Glorot, X., Bordes, A., Bengio, Y.: Deep sparse rectifier neural networks. In:
  International Conference on Artificial Intelligence and Statistics. pp.
  315--323 (2011)

\bibitem{Kang12}
Kang, K.W., Chang, H.J., Shim, H., Kim, Y.J., Choi, B.W., Yang, W.I., Shim,
  J.Y., Ha, J., Chung, N.: Feasibility of an automatic computer-assisted
  algorithm for the detection of significant coronary artery disease in
  patients presenting with acute chest pain. European Journal of Radiology
  81(4),  e640 -- e646 (2012),
  \url{http://www.sciencedirect.com/science/article/pii/S0720048X12000332}

\bibitem{Kiri10}
Kiri{\c{s}}li, H., Schaap, M., Klein, S., Papadopoulou, S., Bonardi, M., Chen,
  C.H., Weustink, A., Mollet, N., Vonken, E., van~der Geest, R., et~al.:
  Evaluation of a multi-atlas based method for segmentation of cardiac {CTA}
  data: a large-scale, multicenter, and multivendor study. Medical Physics
  37(12),  6279--6291 (2010)

\bibitem{Kriz12}
Krizhevsky, A., Sutskever, I., Hinton, G.E.: {ImageNet} classification with
  deep convolutional neural networks. In: Advances in neural information
  processing systems. pp. 1097--1105 (2012)

\bibitem{Lecu15}
LeCun, Y., Bengio, Y., Hinton, G.: Deep learning. Nature  521(7553),  436--444
  (2015)

\bibitem{MP06}
Marie-Pierre, J.: Automatic segmentation of the left ventricle in cardiac {MR}
  and {CT} images. International Journal of Computer Vision  70(2),  151--163
  (2006)

\bibitem{Sriv14}
Srivastava, N., Hinton, G., Krizhevsky, A., Sutskever, I., Salakhutdinov, R.:
  Dropout: A simple way to prevent neural networks from overfitting. The
  Journal of Machine Learning Research  15(1),  1929--1958 (2014)

\bibitem{Suge06}
Sugeng, L., Mor-Avi, V., Weinert, L., Niel, J., Ebner, C.,
  Steringer-Mascherbauer, R., Schmidt, F., Galuschky, C., Schummers, G., Lang,
  R.M., et~al.: Quantitative assessment of left ventricular size and function
  side-by-side comparison of real-time three-dimensional echocardiography and
  computed tomography with magnetic resonance reference. Circulation  114(7),
  654--661 (2006)

\bibitem{Tava13}
Tavakoli, V., Amini, A.A.: A survey of shaped-based registration and
  segmentation techniques for cardiac images. Computer Vision and Image
  Understanding  117(9),  966--989 (2013)

\bibitem{Tech11}
Techasith, T., Cury, R.C.: Stress myocardial {CT} perfusion: an update and
  future perspective. JACC: Cardiovascular Imaging  4(8),  905--916 (2011)

\bibitem{Vos16}
de~Vos, B.D., Wolterink, J.M., de~Jong, P.A., Viergever, M.A., I{\v{s}}gum, I.:
  {2D} image classification for {3D} anatomy localization: employing deep
  convolutional neural networks. In: SPIE Medical Imaging. pp. 97841Y--97841Y
  (2016)

\bibitem{Wolt15}
Wolterink, J.M., Leiner, T., Viergever, M.A., I{\v{s}}gum, I.: Automatic
  coronary calcium scoring in cardiac {CT} angiography using convolutional
  neural networks. In: Medical Image Computing and Computer-Assisted
  Intervention--MICCAI 2015, pp. 589--596. Springer (2015)

\bibitem{WHO14}
{World Health Organization}: Global status report on noncommunicable diseases
  2014 (2014)

\bibitem{Xion15}
Xiong, G., Kola, D., Heo, R., Elmore, K., Cho, I., Min, J.K.: Myocardial
  perfusion analysis in cardiac computed tomography angiographic images at
  rest. Medical Image Analysis  24(1),  77--89 (2015)

\bibitem{Zhen08a}
Zheng, Y., Barbu, A., Georgescu, B., Scheuering, M., Comaniciu, D.:
  Four-chamber heart modeling and automatic segmentation for {3-D} cardiac {CT}
  volumes using marginal space learning and steerable features. IEEE
  Transactions on Medical Imaging  27(11),  1668--1681 (2008)

\end{thebibliography}
\end{document}